# Monte Carlo inference via greedy importance sampling

**Dale Schuurmans** and **Finnegan Southey**
Department of Computer Science
University of Waterloo
{dale,fdjsouth}@cs.uwaterloo.ca

## Abstract

We present a new method for conducting Monte Carlo inference in graphical models which combines explicit search with generalized importance sampling. The idea is to reduce the variance of importance sampling by searching for significant points in the target distribution. We prove that it is possible to introduce search and still maintain unbiasedness. We then demonstrate our procedure on a few simple inference tasks and show that it can improve the inference quality of standard MCMC methods, including Gibbs sampling, Metropolis sampling, and Hybrid Monte Carlo. This paper extends previous work which showed how greedy importance sampling could be correctly realized in the one-dimensional case.

## 1 Introduction

It is well known that general inference and learning with Bayesian networks is computationally hard [DL93, Rot93], and it is therefore necessary to consider restricted architectures [Pea88], or heuristic and approximate algorithms to perform these tasks [JGJS98, Fre98, DL97]. Among the most convenient and successful techniques are stochastic methods which are guaranteed to converge to a correct solution in the limit of large random samples [Mac98, Nea96, Tan93, SP90, Gew89]. These methods can be easily applied to complex inference problems that overwhelm deterministic approaches.

The family of stochastic inference methods can be grouped into the *independent* Monte Carlo methods (importance sampling and rejection sampling [Mac98, FD94, SP90, Gew89, Hen88]) and the *dependent* Markov Chain Monte Carlo (MCMC) methods (Gibbs sampling, Metropolis sampling, and Hybrid Monte Carlo) [Mac98, GRS96, Nea93, Tan93]. The goal of all these methods is to simulate drawing a random sample from a target distribution

$P(x)$ (generally defined by a Bayesian network or graphical model) that is hard to sample from directly.

In this paper we investigate a simple modification of importance sampling that demonstrates some advantages over independent and dependent-Markov-chain methods. The idea is to explicitly search for important regions in a target distribution $P$ when sampling from a simpler proposal distribution $Q$. Previous work [Sch99] showed that search could be incorporated in an importance sampler while still maintaining unbiasedness. However, these results were primarily restricted to the one-dimensional case. In this paper we extend these results by introducing a new estimator that can be successfully applied to multidimensional problems, as well as continuous problems.

The motivation behind our approach is simple: Importance sampling simulates drawing a sample from a target distribution $P$ by drawing a sample from a simpler "proposal" distribution $Q$ and then weighting the sample points $x_1, ..., x_t$ according to $w(x_i) = P(x_i)/Q(x_i)$ to ensure that the expected weight of any point $x$ is $P(x)$. The technique is effective when $Q$ approximates $P$ over most of the domain. However, importance sampling fails when $Q$ misses high probability regions of $P$ and systematically yields sample points with small weights. This situation is illustrated in Figure 1. In these cases, the sample will almost always consist of low weight points, but with small probability it will contain some very high weight points— and this combination causes any estimator based on these samples to have high variance (since the effective sample size is reduced). It is therefore critical to obtain sample points from the important regions of $P$. MCMC methods such as the Metropolis algorithm and Hybrid Monte Carlo attempt to overcome this difficulty by biasing a local random search towards higher probability regions while preserving the asymptotic "fair sampling" properties of the techniques [Nea93, Nea96].

Here we investigate a simpler direct approach where one draws points from a proposal distribution $Q$ but then explicitly searches in $P$ to find points in important regions. The main challenge is to maintain correctness (*i.e.*, unbi-asedness) of the resulting procedure. We achieve this by initiating independent search paths from different starting points and then weighting all of the sample points so that their expected weight under the sampling procedure matches their true probability under the target $P$ (or is at least proportional to this probability; see below). For example, if a point $x$ appears in a sample path with some probability $Q^*(x)$ (which is a function of the proposal distribution $Q$, but is not necessarily just $Q(x)$) then we need to assign the weight $w(x) = P(x)/Q^*(x)$ so that $x$'s expected weight is $P(x)$. By conducting a number of searches, the idea is to build a sample from independent blocks that are each guaranteed to contain high weight points. In this way a reasonable sample of independent high weight points can be captured, which should yield improved inference.

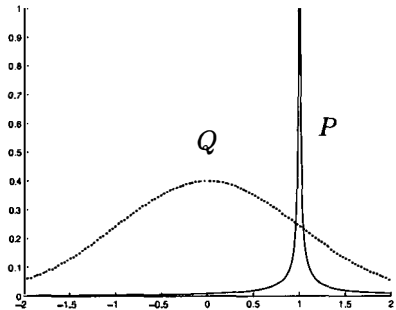

Figure 1: Difficult situation for importance sampling

The remainder of this paper is organized as follows. We first consider the discrete case. Section 2 reviews the basic importance sampling procedure and its application to inference in graphical models. Section 3 then presents a generalization of importance sampling that forms the core of our discrete estimators. We prove that this generalization preserves unbiasedness while admitting new estimation procedures that had not been previously investigated. In Section 4 we then introduce the basic greedy search method we propose in this paper, and prove that it yields unbiased estimates, even in the multidimensional case. Section 5 presents an experimental evaluation of the basic method and compares its performance against standard Monte Carlo inference methods. We then consider the continuous case. Section 6 extends our method to the continuous setting and extends the proof of unbiasedness (which involves some technical complications that are discussed in the appendix). Section 7 then presents an experimental evaluation of the continuous extension.

## 2  Importance sampling

Many inference problems in graphical models can be cast as determining the expected value of a random variable of interest, $f$, given observations drawn according to a target distribution $P$. That is, we are interested in computing

**Importance sampling**

- Draw $x_1, ..., x_n$ independently according to $Q$.
- Weight each sample point $x_i$ by $w(x_i) = \frac{P(x_i)}{Q(x_i)}$.
- For a random variable of interest, $f$, estimate
  $E_{P(x)}f(x)$ by    $\hat{f} = \frac{1}{n}\sum_{i=1}^{n} f(x_i)w(x_i)$.

Figure 2: Importance sampling procedure

$E_{P(x)}f(x)$. Usually the random variable $f$ is simple, like the indicator of some event, but the distribution $P$ is usually not in a form that we can sample from efficiently.

Importance sampling is a useful technique for estimating $E_{P(x)}f(x)$ when $P$ cannot be sampled from directly. The idea is to draw independent points $x_1, ..., x_n$ according to a simpler proposal distribution $Q$, which can be sampled from efficiently, but then weight these points according to $w(x) = P(x)/Q(x)$. Assuming that we can evaluate $P(x)$ at individual sample points, the weighted sample can be computed and used to estimate desired expectations; as shown in Figure 2. The correctness (*i.e.*, unbiasedness) of this procedure is easy to establish. Given a target random variable $f$ with the expected value under $P$ of $E_{P(x)}f(x) = \sum_{x \in X} f(x)P(x)$, we can determine that the expected *weighted* value of $f$ under $Q$ is $E_{Q(x)}f(x)w(x) = \sum_{x \in X} f(x)w(x)Q(x) = \sum_{x \in X} f(x)\frac{P(x)}{Q(x)}Q(x) = \sum_{x \in X} f(x)P(x) = E_{P(x)}f(x)$, thus yielding an unbiased estimate.[1]

Unfortunately, for standard inference problems in Bayesian networks it is usually not possible to implement this procedure directly. The problem is that assigning the weights requires the ability to evaluate $P(x)$ at chosen points $x$, and in Bayesian networks this probability usually corresponds to $P_{BN}(\mathbf{x}|\mathbf{e})$ for a vector of observation variables $\mathbf{x}$ given evidence variables $\mathbf{e}$. In general, it is not possible to efficiently compute [Coo90] or even approximate [Rot93, DL93] these quantities. However, it is possible to apply a generalized algorithm to attempt inference in these cases. Figure 3 shows the "indirect importance sampling" procedure, which does not assign sample weights in the desired form, but rather assigns indirect weights $u(x)$ that are only a fixed constant multiple $\beta$ of the desired ratio, *i.e.*, $u(x) = \beta P(x)/Q(x)$, where $\beta$ need not be known *a priori*. Here we only need to be able to compute $u(x)$ without having to determine $\beta$ directly. This relaxation permits efficient calculation of the desired weights in Bayesian network inference, and results in a procedure that is known as the "likelihood weighting" algorithm [SP90]. For example, when evidence is strictly at the leaves, the likelihood weighting proposal distribution corresponds to the prior $Q(\mathbf{x}) = P_{BN}(\mathbf{x})$ and the indirect weights are given by $u(\mathbf{x}) = P_{BN}(\mathbf{e}|\mathbf{x})$, which is easily computed from

---

[1]We find it advantageous to first present the discrete case for the purposes of clarity. We consider the continuous case later.

**"Indirect" importance sampling** (likelihood weighting)

- Draw $x_1, ..., x_n$ independently according to $Q$.
- Weight each sample point $x_i$ by $u(x_i) = \beta \frac{P(x_i)}{Q(x_i)}$.
- For a random variable of interest, $f$, estimate

$$\mathrm{E}_{P(x)} f(x) \text{ by } \quad \hat{f} = \frac{\sum_{i=1}^n f(x_i) u(x_i)}{\sum_{i=1}^n u(x_i)}.$$

Figure 3: Indirect importance sampling procedure

**x.** These weights $u(x) = P_{BN}(\mathbf{x}|\mathbf{e}) P_{BN}(\mathbf{e}) / P_{BN}(\mathbf{x}) = P_{BN}(\mathbf{e}) w(\mathbf{x})$ are a fixed constant multiple of the desired weights (which *cannot* be computed efficiently), and therefore we can use them to feasibly implement the indirect importance sampling procedure of Figure 3.

The drawback of the indirect procedure is that it is not unbiased at small sample sizes. Rather it only becomes unbiased in the large sample limit. To establish the asymptotic correctness of indirect importance sampling, note that we estimate $\mathrm{E}_{P(x)} f(x)$ by the weighted average of $f$ divided by the sum of indirect weights (Figure 3). It is easy to prove this estimate becomes unbiased because the two weighted averages $\frac{1}{n} \sum_{i=1}^n f(x_i) u(x_i)$ and $\frac{1}{n} \sum_{i=1}^n u(x_i)$ converge to $\beta \mathrm{E}_{P(x)} f(x)$ and $\beta$ respectively, and thus $\hat{f} \to \mathrm{E}_{P(x)} f(x)$ (under broad technical conditions [Gew89]). For the extensions we discuss below it will always be possible to generalize them in a similar manner so that they can be applied to Bayesian network inference. However, to keep the presentation simple we will focus on the simple version of importance sampling described in Figure 2, and establish unbiasedness for that case.

## 3 Generalized importance sampling

As previously discussed, importance sampling is an effective estimation technique when $Q$ approximates $P$ over most of the domain, but it fails when $Q$ misses high probability regions of $P$ (and therefore systematically yields samples with small weights). This problem can occur in Bayesian network inference whenever the evidence is unlikely under the prior distribution, because this causes likelihood weighting to use a proposal distribution that systematically generates points with small weights. For example, when evidence is strictly at the leaves, likelihood weighting uses the proposal distribution $Q(\mathbf{x}) = P_{BN}(\mathbf{x})$, which guarantees that the indirect weights $u(\mathbf{x}) = P_{BN}(\mathbf{e}|\mathbf{x})$ will almost always be small. To overcome this problem it is critical to obtain data points from the important regions of $P$. Our goal is to avoid generating systematically underweight samples by explicitly *searching* for significant regions in the target distribution $P$. To do this, and maintain the unbiasedness of the resulting procedure, we develop a simple generalization of importance sampling that can be proved correct.

Consider a procedure where instead of sampling individ-

**"Generalized" importance sampling**

- Draw $x_1, ..., x_n$ independently according to $Q$.
- For each $x_i$, recover its block $B_i = \{x_{i,1}, ..., x_{i,b_i}\}$.
- Create a large sample out of the blocks

$$x_{1,1}, ..., x_{1,b_1}, x_{2,1}, ..., x_{2,b_2}, ..., x_{n,1}, ..., x_{n,b_n}.$$

- Weight each point $x_j \in B_i$ by $w_i(x_j) = \frac{P(x_j)}{Q(x_i)} \alpha_{ij}$
- For a random variable of interest, $f$, estimate

$$\mathrm{E}_{p(x)} f(x) \text{ by } \quad \hat{f} = \frac{1}{n} \sum_{i=1}^n \sum_{k=1}^{b_i} f(x_{i,k}) w_i(x_{i,k}).$$

Figure 4: Generalized importance sampling procedure

ual points we sample deterministic *blocks* of points. That is, to each domain point $x_i$ associate a fixed block $B_i = \{x_{i,1}, ..., x_{i,b_i}\}$ *a priori*. When $x_i$ is drawn from the proposal distribution $Q$ we use it to recover the block $B_i$ and then add these points to the sample. (We make no restriction on the structure of the blocks, other than they be finite—blocks can overlap and need not contain their initiating $x_i$—however we do require that each domain point $x_i$ deterministically generate the same block $B_i$.) In this scheme, the final sample is built out of independently sampled blocks of points.

The real issue is how to weight the points. To do this we introduce an auxiliary weighting scheme. For each pair of points $x_i, x_j$ define the indicator $I_{ij}$ such that $I_{ij} = 1$ if $x_j \in B_i$ and $I_{ij} = 0$ if $x_j \notin B_i$. Then for every point $x_j$ appearing in a block $B_i$ (initiated by $x_i$) we associate a nonnegative weight $\alpha_{ij}$. (Here we think of $x_i$ as the initiating point and $x_j$ as one of the successors in its block.) The $\alpha_{ij}$ weights can be more or less arbitrary except that they must be nonnegative and satisfy the constraint

$$\sum_{x_i \in X} \alpha_{ij} I_{ij} = 1 \quad (1)$$

for every $x_j$. That is, for each destination point $x_j$, the total of the incoming $\alpha$-weight has to sum to 1. (We will see below how this can easily be arranged in real procedures.) Note that a particular point $x_j$ might appear in several blocks, but this will be of no concern as long as the above constraint is satisfied. To compute the final weight of a point $x_j$ in a sampled block $B_i$ we use $w_i(x_j) = \frac{P(x_j)}{Q(x_i)} \alpha_{ij}$.

In summary, the generalized importance sampling procedure (Figure 4) draws initial points $x_i$ from $Q$, builds a sample out of the corresponding blocks $B_i$, and weight the individual points $x_j \in B_i$ by $w_i(x_j) = \frac{P(x_j)}{Q(x_i)} \alpha_{ij}$. Although only loosely constrained in terms of the block structure and auxiliary weighting scheme, it turns out that this procedure always yields an unbiased estimate of $\mathrm{E}_{p(x)} f(x)$ for any random variable of interest, $f$. To prove this, consider the expected *weighted* value of $f$ when sampling initiating points $x_i$ under $Q$

$$\mathrm{E}_{Q(x_i)} \left[ \sum_{x_j \in B_i} f(x_j) w_i(x_j) \right]$$

$$\begin{aligned}
&= \sum_{x_i \in X} \sum_{x_j \in B_i} f(x_j) w_i(x_j) Q(x_i) \\
&= \sum_{x_i \in X} \sum_{x_j \in B_i} f(x_j) \frac{P(x_j)}{Q(x_i)} \alpha_{ij} Q(x_i) \\
&= \sum_{x_i \in X} \sum_{x_j \in B_i} f(x_j) P(x_j) \alpha_{ij} \\
&= \sum_{x_i \in X} \sum_{x_j \in X} I_{ij} f(x_j) P(x_j) \alpha_{ij} \\
&= \sum_{x_j \in X} \sum_{x_i \in X} I_{ij} f(x_j) P(x_j) \alpha_{ij} \\
&= \sum_{x_j \in X} f(x_j) P(x_j) \sum_{x_i \in X} I_{ij} \alpha_{ij} \\
&= \sum_{x_j \in X} f(x_j) P(x_j) \;=\; E_{P(x)} f(x)
\end{aligned}$$

thus yielding an unbiased estimate.

This proof is remarkably simple and yet has general implications. One important property of this argument is that it does not depend on *how* the block decomposition is chosen or how the auxiliary weights are set, so long as they satisfy the above constraints. That is, we could fix any block decomposition and auxiliary weighting scheme, even ones that depended on the target distribution $P$, without affecting the correctness of the procedure. Intuitively, this holds because once the block structure and weighting scheme are established, unbiasedness is achieved by randomly sampling blocks and assigning fair weights to the points, regardless of the set up. (Of course, we need the blocks to be finite to be able to implement the procedure.)

The generality of this outcome allows us to contemplate alternative sampling procedures that explicitly attempt to construct blocks that have high weight points. We can then use the sampling procedure outlined in Figure 4 to yield unbiased estimates for any random variable $f$.

## 4 Greedy importance sampling

It is well known that the optimal proposal distribution for regular importance sampling is just $Q^*(x) = |f(x)P(x)|/\sum_{x \in X} |f(x)P(x)|$ when estimating an expectation $E_{P(x)} f(x)$ (which minimizes the variance of the estimate [Eva91, Rub81]). In searching for significant regions in the domain it would appear that one should seek points that have a high value of $|f(x)P(x)|$ not just $P(x)$. Our greedy search procedures, therefore, search for points that maximally increase the objective $|f(x)P(x)|$. In the discrete case we examine a set of immediate neighbors and take a greedy step.

The "greedy" importance sampling procedure outlined in Figure 5 first draws an initial point $x_1$ from $Q$ and then conducts a greedy search in the direction of maximum

---

**"Greedy" importance sampling**
- Draw $x_1, ..., x_n$ independently from $Q$.
- For each $x_i$, let $x_{i,1} = x_i$:
- Compute block $B_i = \{x_{i,1}, x_{i,2}, ..., x_{i,m}\}$ by taking local search steps in the direction of maximum $|f(x)P(x)|$ until a local maximum or $m-1$ steps.
- Weight each point $x_j \in B_i$ by $w_i(x_j) = \frac{P(x_j)}{Q(x_i)} \alpha_{ij}$ where $\alpha_{ij}$ is defined in (2) below.
- Create the final sample from the blocks of points
  $x_{1,1}, ..., x_{1,m}, x_{2,1}, ..., x_{2,m}, ..., x_{n,1}, ..., x_{n,m}.$
- For a random variable, $f$, estimate $E_{P(x)} f(x)$
  by   $\hat{f} = \frac{1}{n} \sum_{i=1}^{n} \sum_{k=1}^{m} f(x_{i,k}) w_i(x_{i,k}).$

Figure 5: "Greedy" importance sampling procedure

---

$|f(x)P(x)|$ among local neighbors, until either $m-1$ steps have been taken or a local maximum is encountered. (To break ties we employ a deterministic policy that prevents loops.) A single "block" in the final sample is comprised of a complete sequence captured in one ascending search.

This procedure is easy to implement in principle. The only challenge is finding ways to assign the auxiliary weights $\alpha_{ij}$ so that they satisfy the constraint (1). That is, we require that the total incoming $\alpha$-weight to any domain point $x_j$ be exactly 1. Note that, in principle, to verify (1) for a domain point $x_j$ we have to consider *every* search path that starts at some other point $x_i$ and passes through $x_j$ within $m-1$ steps. If our search is deterministic (which we assume) the set of search paths entering $x_j$ will form a tree in general. Let $T_j$ denote the tree of points that lead to $x_j$ and let $\alpha(T_j) = \sum_{x_k \in T_j} \alpha_{kj}$. This tree will have depth at most $m$ given the bound on search length. It turns out that it is easy to assign $\alpha$-weights to ensure $\alpha(T_j) = 1$ using the following recursive scheme.

Fix a parameter $b$ which serves as a guess of the typical "inward" branching factor of trees in the domain. A complete balanced tree of depth $m$ with inward branching factor $b$ will have a number of nodes equal to

$$S_{b,m} = 1 + b + b^2 + \cdots + b^{m-1} = \begin{cases} \frac{b^m - 1}{b - 1} & \text{if } b \neq 1 \\ m & \text{if } b = 1 \end{cases}$$

If $T_j$ were a complete and balanced tree then we could assign a weight $\alpha_{ij} = 1/S_{b,m}$ for every $x_i$ in $T_j$ and trivially satisfy the constraint (1). However, $T_j$ need not be balanced and its internal nodes might have different branching factors. Moreover, we must be able to compute the weights on-line when conducting a search from a single initial point $x_i$. Thus, to assign the $\alpha$-weights we employ the following local correction scheme: Given a start node $x_i$ and a search path $x_i, x_{i+1}, ..., x_{i+k} = x_j$ from $x_i$ to $x_j$, we assign an auxiliary weight $\alpha_{ij}$ by

$$\alpha_{ij} = \begin{cases} \beta_{ij} \frac{1}{S_{b,m}} & \text{if } b_i \neq 0 \\ \beta_{ij} \frac{S_{b,m-k+2}}{S_{b,m}} & \text{if } b_i = 0 \end{cases} \qquad (2)$$

$$\beta_{ij} = \begin{cases} 1 & \text{if } i = j \\ \frac{b}{b_{i+1}} \frac{b}{b_{i+2}} \cdots \frac{b}{b_{i+k}} & \text{o.w. (where } x_{i+k} = x_j) \end{cases}$$

where $b_{i+\ell}$ denotes the inward branching factor of node $x_{i+\ell}$. This scheme reflects the principle that a node $x_i$ at a distance $k$ from the root of $T_j$ takes responsibility for a complete balanced subtree of height $m - (k-1)$ below $x_i$.[2]

It is easy to show that assigning auxiliary weights in this way preserves the desired constraint $\alpha(T_j) = 1$. The proof is by an induction on the level of the subtrees rooted at internal nodes $x_i \in T_j$: Let $T_{ij}$ denote the subtree of $T_j$ rooted at $x_i$. We establish that for subtrees at levels $\ell = 1, 2, ..., m$ (moving from the leaves up) that our system of auxiliary weights satisfies the invariant

$$\alpha(T_{ij}) = \beta_{ij} \frac{S_{b,\ell}}{S_{b,m}} \qquad (3)$$

Once this is established, it will trivially follow that $\alpha(T_j) = \alpha(T_{jj}) = \beta_{jj} S_{b,m} / S_{b,m} = 1$.

For the base case note that for a subtree $T_{ij}$ at level 1 (i.e., a leaf) we have $T_{ij} = \{x_i\}$ and hence $\alpha(T_{ij}) = \alpha_{ij} = \beta_{ij} / S_{b,m} = \beta_{ij} S_{b,1} / S_{b,m}$, which satisfies (3). Now assume as an induction hypothesis that (3) holds for trees of level $\ell - 1$ in $T_j$; we prove that (3) must then hold for all trees of level $\ell$. Consider a node $x_i$ at level $\ell$ in $T_j$ and assume $b_i \neq 0$. (The proof for the case when $b_i = 0$ is similar; we omit it for brevity.) In this situation we have

$$
\begin{aligned}
\alpha(T_{ij}) &= \alpha_{ij} + \sum_{x_k : x_k \overset{\text{immed}}{\to} x_i} \alpha(T_{kj}) \\
&= \frac{\beta_{ij}}{S_{b,m}} + \sum_{x_k : x_k \overset{\text{immed}}{\to} x_i} \beta_{kj} \frac{S_{b,\ell-1}}{S_{b,m}} \\
&\quad \text{by (2) and the induction hypothesis} \\
&= \frac{\beta_{ij}}{S_{b,m}} + \sum_{x_k : x_k \overset{\text{immed}}{\to} x_i} \beta_{ij} \frac{b}{b_i} \frac{S_{b,\ell-1}}{S_{b,m}} \\
&\quad \text{by the definition of } \beta_{ij} \\
&= \frac{\beta_{ij}}{S_{b,m}} + b_i \left( \beta_{ij} \frac{b}{b_i} \frac{S_{b,\ell-1}}{S_{b,m}} \right) \\
&= \beta_{ij} (1 + b S_{b,\ell-1}) / S_{b,m} \\
&= \beta_{ij} S_{b,\ell} / S_{b,m} \\
&\quad \text{by the definition of } S_{b,\ell}
\end{aligned}
$$

---

[2]The idea behind the auxiliary weighting scheme is that if the node weights all start out being 1 and $x_i$'s branching factor $b_i$ differs from $b$, then $x_i$ will compensate by multiplying the weights of its subtrees by $b/b_i$ so that the total node weight of the subtree rooted at $x_i$ remains $1 + \frac{b}{b_i}(b_i S_{b,m-(k-2)}) = 1 + b S_{b,m-(k-2)} = S_{b,m-(k-1)}$. In the case when $b_i = 0$, $x_i$ will compensate by adding the equivalent of $b$ subtrees of depth $m - (k-2)$ to its own node weight, so that again $1 + b S_{b,m-(k-2)} = S_{b,m-(k-1)}$. In either case, as far as the rest of the tree $T_j$ is concerned, $x_i$ will be the root of a complete balanced subtree of height $m - (k-1)$.

| $t$ | | DS | GIS | IS | Met |
|---|---|---|---|---|---|
| 100 | bias | 0.0002 | 0.0159 | 0.0848 | 15.6036 |
| | stdev | 0.1003 | 0.1871 | 0.3477 | 13.9486 |
| | rmse | 0.1003 | 0.1877 | 0.3579 | 20.9293 |
| 200 | bias | 0.0013 | 0.0013 | 0.0410 | 18.5112 |
| | stdev | 0.0696 | 0.1303 | 0.2183 | 13.4202 |
| | rmse | 0.0696 | 0.1303 | 0.2221 | 22.8641 |
| 500 | bias | 0.0007 | 0.0013 | 0.0160 | 21.7949 |
| | stdev | 0.0439 | 0.0819 | 0.1324 | 12.7147 |
| | rmse | 0.0439 | 0.0819 | 0.1333 | 25.2326 |
| 700 | bias | 0.0014 | 0.0040 | 0.0110 | 24.7202 |
| | stdev | 0.0375 | 0.0728 | 0.1135 | 12.2979 |
| | rmse | 0.0376 | 0.0729 | 0.1140 | 27.6103 |
| 1000 | bias | 0.0020 | 0.0014 | 0.0097 | 26.1688 |
| | stdev | 0.0315 | 0.0609 | 0.0860 | 11.9118 |
| | rmse | 0.0316 | 0.0609 | 0.0865 | 28.7524 |

Table 1: Discrete two-dimensional experiments: Scaling in estimation sample size $t$. Distributions are bivariate Gaussians discretized on 21 x 21 grid. 1000 repetitions. $\mu_P = \mu_Q = 0$, $\Sigma_P = I$, $\Sigma_Q = 6^2 I$, $f(x) = -\log(p(x))$.

which satisfies the desired invariant (3).

Thus, we have a general procedure for conducting an explicit search in general spaces, where search paths can merge, and yet our weighting scheme still correctly compensates and yields unbiased estimates.

## 5 Experiments

To demonstrate the greedy method we conducted a series of experiments in order to gain an understanding of its performance relative to standard techniques. Given the unbiasedness of the candidate methods, the main issue is to assess the variance of the estimators. To measure the performance of the various techniques, we gathered their mean squared error, bias and variance over 1000 repetitions of a given problem. We use root mean squared error (rmse), absolute bias, and standard deviation to give an intuitive measure of an estimator's effectiveness. The methods we compared were greedy importance sampling (GIS), direct sampling from the target distribution (DS), standard importance sampling (IS), rejection sampling (RS), and Metropolis sampling [Nea93, Nea96]. For Metropolis sampling we used a uniform neighborhood proposal distribution.

The first series of experiments we conducted was on simple one-dimensional problems that varied the relationship between $P$, $Q$ and $f$, in order to test the basic correctness of our method. Figure 6 shows that GIS strongly outperforms the other methods across a range of problems in this setting. Interestingly, these experiments also show that it is not always best to sample directly from the target distribution $P$ when the random variable $f$ has a substantially different structure. We also tested GIS on a simple multivariate problem, and Table 1 shows that GIS handles multidimensional problems quite well in this simple case.

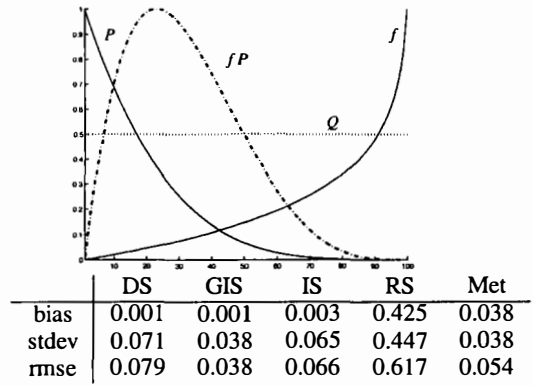

|       | DS    | GIS   | IS    | RS    | Met   |
|-------|-------|-------|-------|-------|-------|
| bias  | 0.001 | 0.001 | 0.003 | 0.425 | 0.038 |
| stdev | 0.071 | 0.038 | 0.065 | 0.447 | 0.038 |
| rmse  | 0.079 | 0.038 | 0.066 | 0.617 | 0.054 |

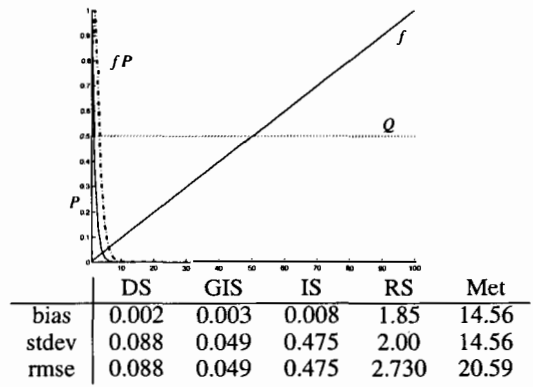

|       | DS    | GIS   | IS    | RS   | Met   |
|-------|-------|-------|-------|------|-------|
| bias  | 0.002 | 0.003 | 0.008 | 1.85 | 14.56 |
| stdev | 0.088 | 0.049 | 0.475 | 2.00 | 14.56 |
| rmse  | 0.088 | 0.049 | 0.475 | 2.730| 20.59 |

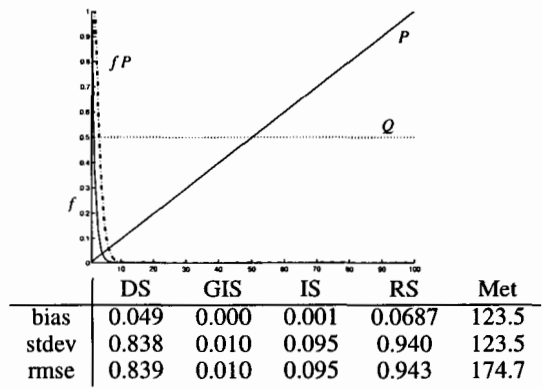

|       | DS    | GIS   | IS    | RS     | Met   |
|-------|-------|-------|-------|--------|-------|
| bias  | 0.049 | 0.000 | 0.001 | 0.0687 | 123.5 |
| stdev | 0.838 | 0.010 | 0.095 | 0.940  | 123.5 |
| rmse  | 0.839 | 0.010 | 0.095 | 0.943  | 174.7 |

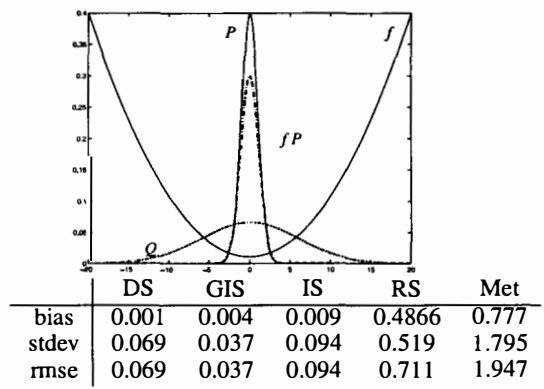

|       | DS    | GIS   | IS    | RS     | Met   |
|-------|-------|-------|-------|--------|-------|
| bias  | 0.001 | 0.004 | 0.009 | 0.4866 | 0.777 |
| stdev | 0.069 | 0.037 | 0.094 | 0.519  | 1.795 |
| rmse  | 0.069 | 0.037 | 0.094 | 0.711  | 1.947 |

Figure 6: Discrete one-dimensional experiments: Problems with varying relationships between $P$, $Q$, $f$ and $|fP|$. 1000 repetitions with estimation samples of size 100.

**Continuous greedy importance sampling**

- Draw $x_1, ..., x_n$ independently from $Q$.
- For each $x_i$, let $x_{i,1} = x_i$:
- Compute block $B_i = \{x_{i,1}, x_{i,2}, ..., x_{i,m}\}$ by taking size $\epsilon$ steps in axis direction of increasing $|f(x)P(x)|$ until a local maximum or $m - 1$ steps have been reached.
- Weight each point $x_j \in B_i$ by $w_i(x_j) = \frac{P(x_j)}{Q(x_j)} \alpha_{ij}$ where $\alpha_{ij}$ is defined in (2) above.
- Create the final sample from the blocks of points
$$x_{1,1}, ..., x_{1,m}, x_{2,1}, ..., x_{2,m}, ..., x_{n,1}, ..., x_{n,m}.$$
- For a random variable, $f$, estimate $E_{P(x)}f(x)$
by $\hat{f} = \frac{1}{n} \sum_{i=1}^{n} \sum_{k=1}^{m} f(x_{i,k}) w_i(x_{i,k})$.

Figure 7: Continuous greedy importance sampling

## 6 Continuous case

Although the greedy importance sampling technique shows promise for simple discrete problems, it is also important to handle continuous problems defined over the sample space $X = \mathbb{R}^n$. Here, regular importance sampling applies just as well as in the discrete case, using weights determined by the ratio of the probability densities, $w(x) = p(x)/q(x)$ (provided the densities exist); see Figure 7. The unbiasedness of this procedure is easy to establish when the densities exist, since $E_{Q(x)}f(x)w(x) = \int_{\mathbb{R}^n} f(x)w(x)q(x) \ \mu(dx) = \int_{\mathbb{R}^n} f(x)\frac{p(x)}{q(x)}q(x) \ \mu(dx) = \int_{\mathbb{R}^n} f(x)p(x) \ \mu(dx) = E_{P(x)}f(x)$. (Note that we use $\mu$ to denote standard Lebesgue measure on $\mathbb{R}^n$.)

Extending the greedy importance sampling strategy to the continuous case is not hard in principle—however, care must be taken not to "warp" the measure on the sample space or else bias could be introduced into the procedure. To avoid the need for compressing or dilating transformations (by determining Jacobians) we employ a very simple search scheme that takes only fixed size steps (of size $\epsilon$) in axis parallel directions. That is, in $n$-dimensional Euclidean space, the greedy importance sampling procedure given in Figure 7 first draws an initial point $x_i$ from $q(x)$ and then conducts a search in the direction of maximum $|f(x)p(x)|$ among the $2n$ neighbors of $x_i$ determined by taking steps of size $\epsilon$ in each direction from $x_i$; until either $m - 1$ steps have been taken or a local maximum is reached. A single "block" in the final sample is comprised of a complete sequence in one ascending search. The remainder of the procedure (computing the $\alpha$-weights) is implemented as in the discrete case (with the $\alpha_{ij}$'s satisfying the same constraints). To break ties, we employ a deterministic tie breaking policy that prevents loops.

The intuitive significance of this "grid walk" strategy is that it does not transform the underlying Lebesgue measure of the Euclidean sample space. Instead it leaves the resulting discrete "merges" to be compensated by the $\alpha$-weighting scheme. In fact, except for one step, it is actually quite easy to prove that this continuous version of the greedy importance sampling procedure also yields unbiased estimates:

$$
\begin{aligned}
\mathrm{E}_{Q(x_i)} & \left[ \sum_{x_j \in B_i} f(x_j) w_i(x_j) \right] \\
&= \int_{\mathbb{R}^n} \left[ \sum_{x_j \in B_i} f(x_j) \frac{p(x_j)}{q(x_i)} \, \alpha_{ij} \right] q(x_i) \, \mu(dx_i) \\
&= \int_{\mathbb{R}^n} \sum_{x_j \in B_i} f(x_j) p(x_j) \, \alpha_{ij} \, \mu(dx_i) \\
&= \int_{\mathbb{R}^n} \sum_{x_i \in B_j} f(x_j) p(x_j) \, \alpha_{ij} \, \mu(dx_i) \qquad (4) \\
&\quad \text{(see appendix)} \\
&= \int_{\mathbb{R}^n} f(x_j) p(x_j) \sum_{x_i \in B_j} \alpha_{ij} \, \mu(dx_j) \\
&= \int_{\mathbb{R}^n} f(x_j) p(x_j) \, \mu(dx_j) \quad = \quad \mathrm{E}_{P(x)} f(x)
\end{aligned}
$$

The only significant step in this derivation (4; changing the order of integration) requires an argument that depends on measure theoretic details, and is given in the appendix.

Thus, we have a general procedure for conducting explicit searches in continuous spaces (in a controlled fashion) that yields unbiased estimates.

# 7 Continuous experiments

To verify that our continuous greedy importance sampler is in fact correct, we ran a series of simple experiments to determine that the bias was zero (up to sampling error). Figure 2 demonstrates this both for a one dimensional and three dimensional problem. (In fact, these results are for the *indirect* versions of the importance sampling procedures, which technically are biased since they use the estimator given in Figure 3. However, our results show that this bias can quickly become negligible, and we have found that the indirect estimator significantly reduces variance for both regular and greedy importance sampling. Therefore we report only these results here.)

To gain an understanding of how effective greedy importance sampling performs relative to state of the art techniques on continuous problems, we then conducted a series of simple experiments meant to challenge standard methods. Again, given the (relative) unbiasedness of the candidate methods, the main issue is to assess the variance of the estimators. To measure the performance of the various techniques, we report root mean squared error (rmse), absolute bias, and standard deviation over 1000 repetitions of a given problem. The methods we compared here were greedy importance sampling (GIS), direct sampling from the target distribution (DS), standard importance sampling (IS), Gibbs sampling, Metropolis sampling (Met), and Hybrid Monte Carlo (HMC) [Nea93, Nea96]. For Metropolis sampling we used a Gaussian proposal distribution with covariance $\Sigma_{met} = I/2$, and for GIS we set the step-size $\epsilon = 1$, walk-length $m = 10n$, and $b = n/2.6$, where $n$ is the dimension of the problem.

|  |  | DS | GIS | IS | Met | HMC |
|---|---|---|---|---|---|---|
| $n = 1$ |  |  |  |  |  |  |
| $t = 100$ | bias | 0.003 | 0.003 | 0.010 | 0.197 | 0.292 |
|  | stdev | 0.068 | 0.052 | 0.092 | 1.068 | 0.249 |
|  | rmse | 0.068 | 0.052 | 0.093 | 1.086 | 0.383 |
| 200 | bias | 0.004 | 0.022 | 0.005 | 0.084 | 0.293 |
|  | stdev | 0.050 | 0.032 | 0.066 | 0.793 | 0.244 |
|  | rmse | 0.050 | 0.039 | 0.066 | 0.798 | 0.381 |
| 500 | bias | 0.001 | 0.001 | 0.002 | 0.028 | 0.293 |
|  | stdev | 0.032 | 0.023 | 0.040 | 0.187 | 0.242 |
|  | rmse | 0.032 | 0.023 | 0.040 | 0.189 | 0.381 |
| 1000 | bias | 0.000 | 0.000 | 0.001 | 0.010 | 0.294 |
|  | stdev | 0.022 | 0.016 | 0.029 | 0.131 | 0.241 |
|  | rmse | 0.022 | 0.016 | 0.029 | 0.131 | 0.380 |
| 2000 | bias | 0.001 | 0.000 | 0.001 | 0.008 | 0.293 |
|  | stdev | 0.016 | 0.011 | 0.021 | 0.072 | 0.242 |
|  | rmse | 0.016 | 0.011 | 0.021 | 0.073 | 0.381 |
| $n = 3$ |  |  |  |  |  |  |
| $t = 100$ | bias | 0.001 | 0.324 | 0.863 | 2.806 | 0.961 |
|  | stdev | 0.118 | 0.539 | 1.319 | 7.031 | 0.313 |
|  | rmse | 0.118 | 0.629 | 1.576 | 7.570 | 1.011 |
| 200 | bias | 0.003 | 0.164 | 0.363 | 0.621 | 0.961 |
|  | stdev | 0.086 | 0.394 | 0.834 | 1.673 | 0.306 |
|  | rmse | 0.086 | 0.427 | 0.909 | 1.784 | 1.009 |
| 500 | bias | 0.002 | 0.056 | 0.113 | 0.296 | 0.962 |
|  | stdev | 0.056 | 0.241 | 0.441 | 0.901 | 0.302 |
|  | rmse | 0.056 | 0.247 | 0.455 | 0.948 | 1.009 |
| 1000 | bias | 0.000 | 0.033 | 0.076 | 0.123 | 0.962 |
|  | stdev | 0.039 | 0.160 | 0.287 | 0.383 | 0.301 |
|  | rmse | 0.039 | 0.163 | 0.297 | 0.402 | 1.008 |
| 2000 | bias | 0.000 | 0.008 | 0.030 | 0.060 | 0.963 |
|  | stdev | 0.028 | 0.111 | 0.204 | 0.189 | 0.300 |
|  | rmse | 0.028 | 0.112 | 0.206 | 0.198 | 1.008 |

Table 2: Continuous experiments: Scaling in estimation sample size $t$. Distributions are Gaussians of dimension $n = 1$ and $n = 3$. 1000 repetitions. $\mu_P = \mu_Q = 0$, $\Sigma_P = I$, $\Sigma_Q = 6^2 I$, $f(x) = -\log(p(x))$.

| $n$ |  | DS | GIS | IS | Met | HMC |
|---|---|---|---|---|---|---|
| 1 | bias | 0.001 | 0.000 | 0.001 | 0.017 | 0.294 |
|  | stdev | 0.023 | 0.016 | 0.029 | 0.121 | 0.241 |
|  | rmse | 0.023 | 0.016 | 0.029 | 0.122 | 0.380 |
| 2 | bias | 0.000 | 0.026 | 0.012 | 0.051 | 0.609 |
|  | stdev | 0.030 | 0.065 | 0.099 | 0.272 | 0.293 |
|  | rmse | 0.030 | 0.070 | 0.100 | 0.277 | 0.676 |
| 3 | bias | 0.000 | 0.033 | 0.062 | 0.125 | 0.962 |
|  | stdev | 0.039 | 0.160 | 0.298 | 0.379 | 0.300 |
|  | rmse | 0.039 | 0.163 | 0.304 | 0.399 | 1.008 |
| 5 | bias | 0.001 | 0.237 | 1.442 | 0.421 | 1.748 |
|  | stdev | 0.050 | 0.373 | 1.614 | 0.956 | 0.268 |
|  | rmse | 0.050 | 0.442 | 2.164 | 1.044 | 1.769 |
| 10 | bias | 0.000 | 0.732 | 18.85 | 2.026 | 1.996 |
|  | stdev | 0.072 | 0.620 | 6.470 | 2.124 | 0.278 |
|  | rmse | 0.072 | 0.959 | 19.93 | 2.936 | 2.016 |
| 15 | bias | 0.001 | 1.019 | 68.80 | 11.82 | 4.008 |
|  | stdev | 0.180 | 0.896 | 15.22 | 12.47 | 0.157 |
|  | rmse | 0.180 | 1.358 | 70.46 | 17.18 | 4.011 |

Table 3: Continuous multi-dimensional experiments: Scaling in dimension $n$. Distributions are multivariate Gaussians. 1000 repetitions with estimation samples of size 1000. $\mu_P = \mu_Q = 0$, $\Sigma_P = I$, $\Sigma_Q = 6^2 I$, $f(x) = -\log(p(x))$.

| | DS | GIS | IS | Met |
|---|---|---|---|---|
| $n = 1$ | rmse | | | |
| $\Sigma_Q = 3^2 I$ | 0.050 | 0.058 | 0.047 | 0.20 |
| $6^2$ | 0.050 | 0.039 | 0.066 | 0.80 |
| $9^2$ | 0.050 | 0.037 | 0.081 | 3.79 |
| $12^2$ | 0.049 | 0.036 | 0.092 | 9.58 |
| $15^2$ | 0.049 | 0.034 | 0.103 | 21.8 |
| $n = 5$ | rmse | | | |
| $\Sigma_Q = 3^2 I$ | 0.111 | 0.579 | 0.85 | 0.63 |
| $6^2$ | 0.112 | 0.441 | 5.15 | 5.01 |
| $9^2$ | 0.110 | 0.499 | 13.5 | 25.1 |
| $12^2$ | 0.106 | 0.911 | 25.1 | 71.1 |
| $15^2$ | 0.109 | 1.140 | 42.0 | 155 |
| $n = 10$ | rmse | | | |
| $\Sigma_Q = 3^2 I$ | 0.153 | 1.46 | 4.98 | 1.84 |
| $6^2$ | 0.147 | 0.88 | 30.5 | 17.2 |
| $9^2$ | 0.150 | 2.50 | 74.0 | 71.4 |
| $12^2$ | 0.142 | 4.66 | 133 | 177 |
| $15^2$ | 0.138 | 6.04 | 210 | 371 |

Table 4: Continuous multi-dimensional experiments: Scaling in dimension $n$ as proposal distribution $Q$ is widened. Distributions are multivariate Gaussians. 1000 repetitions with estimation samples of size 200. $\mu_P = \mu_Q = 0$, $\Sigma_P = I$, $f(x) = -\log(p(x))$.

| $t$ | DS | GIS | IS | Met | Gibbs |
|---|---|---|---|---|---|
| | rmse | | | | |
| 100 | 25 | 241 | 254 | 257 | 322 |
| 200 | 17 | 233 | 253 | 256 | 317 |
| 300 | 15 | 223 | 250 | 256 | 316 |
| 500 | 12 | 213 | 249 | 256 | 315 |
| 700 | 10 | 205 | 247 | 256 | 315 |
| 1000 | 8 | 184 | 243 | 256 | 314 |
| 1500 | 7 | 175 | 238 | 256 | 314 |
| 2000 | 6 | 165 | 235 | 256 | 314 |
| 3000 | 5 | 154 | 225 | 256 | 314 |

Table 5: Mixture of Gaussians experiment: Scaling in sample size $t$. Target distribution is a mixture of two bivariate Gaussians, proposal distribution is a single bivariate Gaussian. 1000 repetitions. $\mu_{P_0} = \mu_Q = [0; 0]$, $\mu_{P_1} = [16; 16]$, $\Sigma_{P_t} = I$, $\Sigma_Q = 6^2 I$.

| | IS | GIS | Particle filter |
|---|---|---|---|
| bias | 2.7731 | 1.0764 | 0.8712 |
| stdev | 1.2107 | 0.1079 | 0.2134 |
| rmse | 3.0259 | 1.0818 | 0.8970 |

Table 6: Dynamic probabilistic inference: Estimated value of final state given first six observations. 500 repetitions.

To test GIS on higher dimensional problems, we investigated a series of multivariate Gaussian problems where the relationship between $P, Q$ and $f$ was varied. Here we used a fixed target function $f(x) = -\log(p(x))$ (and hence were estimating the differential entropy of the target distribution $P$ [CT91]). Table 3 shows that GIS can scale up effectively in the number of dimensions, and seems to handle multidimensional problems quite well in this simple setting. In fact, Table 3 shows that the advantage demonstrated by GIS actually appears to grow with dimensionality.

We also ran a series of experiments that varied the width of the proposal distribution $Q$. Table 4 shows that weakening $Q$ damaged the performance of importance sampling (predictably), while the greedy search seemed to mitigate the effects of a poor $Q$ to the extent that its detrimental effects are significantly diminished.

Next, to scale up to a slightly more complex task we considered a distribution $P$ that was a mixture of Gaussians while keeping $Q$ unimodal. In this case, we once again find that GIS performs reasonably well. Table 5 shows that GIS exhibits good performance on this problem and appears to be converging faster than the other methods as the sample size is increased.

Finally, to attempt a more significant study, we applied GIS to an inference problem in graphical models: recovering the hidden state sequence from a dynamic probabilistic model, given a sequence of observations. Here we considered a simple Kalman filter model which had one state variable and one observation variable per timestep, and used the conditional distributions $X_t | X_{t-1} \sim N(x_{t-1}, \sigma_s^2)$, $Z_t | X_t \sim N(x_t, \sigma_o^2)$ and initial distribution $X_1 \sim N(0, \sigma_s^2)$. The problem was to infer the value of the final state variable $x_t$ given the observations $z_1, z_2, ..., z_t$. Table 6 again demonstrates that GIS has a sizeable advantage over standard importance sampling. (In fact, the greedy approach can be applied to "particle filtering" [IB96, KKR95] to obtain further improvements on this task, but space bounds preclude a detailed discussion.)

## 8 Conclusions

We have introduced a new approach to reducing the variance of importance sampling that is provably unbiased in the multidimensional and continuous cases. Our experimental results, although limited to simple synthetic problems, do suggest the bare plausibility of the approach. It appears that, by capturing a moderate size sample of independent high weight points, greedy importance sampling is able to outperform methods that are unlikely to observe such points by chance. Demonstrating the true effectiveness of the approach on real problems remains to be done.

For future work, an important research issue is to perform a variance analysis of the various procedures to analytically determine their relative advantages. An alternative approach to variance reduction is to use *stratified* sampling [Bou94]. This is an orthogonal approach to the method pursued in this paper, and incorporating stratified sampling in a greedy search scheme remains an interesting direction for future research. Another approach to variance reduction, specific to inference in Bayesian networks, is to modify the proposal distribution and weighting scheme as individual variables are sampled in the network [CHM96]. Incorporating these ideas might also be a fruitful direction.

# A Appendix

Let $B_i$ be the finite set of points visited by the greedy search procedure when starting at point $x_i$; let $B_j$ be the finite set of starting points that reach $x_j$; and let $B$ be the relation $(x_i, x_j) \in B$ iff the greedy search process, starting at $x_i$, reaches $x_j$. We wish to show that for a function $h(x_i, x_j)$ we can switch the order of integration

$$\int_{\mathbb{R}^n} \sum_{x_j \in B_i} h(x_i, x_j) \; \alpha_{ij} \; \mu(dx_i)$$

$$= \int_{\mathbb{R}^n} \sum_{x_i \in B_j} h(x_i, x_j) \; \alpha_{ij} \; \mu(dx_j)$$

where $\mu$ is Lebesgue measure on $\mathbb{R}^n$.

Before addressing the main question, we first consider the underlying measures on the spaces $X_i = \mathbb{R}^n$, $X_j = \mathbb{R}^n$, and $X_i \times X_j = \mathbb{R}^{2n}$. Let $\mu_i(\cdot)$ denote the Lebesgue measure on $X_i$. For each initiating point $x_i$ we have a conditional measure given by

$$\mu_{j|i}(x_j|x_i) = \begin{cases} \alpha_{ij} & \text{if } (x_i, x_j) \in B \\ 0 & \text{otherwise} \end{cases}$$

Note that $\mu_i$ and $\mu_{j|i}$ will define a unique joint measure $\mu_{ij}$ on $X_i \times X_j$ via the (generalized) product measure theorem [Ash72, Theorem 2.6.2] (assuming benign technical conditions on $B$). Although this joint measure might seem a bit peculiar, since $\mu_i$ is Lebesgue measure and $\mu_{j|i}$ is a discrete measure with "impulses" at $x_j \in B_i$ of weight $\alpha_{ij}$, it nevertheless gives a well defined measure on $X_i \times X_j$.

We will now show that starting with a uniform measure $\mu_i$ on initiating points $x_i$ and passing through the weighted search process, the result will still be a uniform measure $\mu_j$ on destination points $x_j$. That is, the marginal measure $\mu_j$ on $X_j$ satisfies $\mu_j(A) = \mu_i(A)$ for all Borel sets $A$, and hence is also Lebesgue measure.

**Lemma 1** *For any hypercube $A \subset X_j$ with edge length less than the step size $\epsilon$, $\mu_j(A) = \mu_i(A)$.*

**Proof** Consider some destination point $x_j \in A$. This point will have a predecessor tree of initiating points that reach $x_j$ via the search mechanism. Partition $A$ into a finite collection of equivalence classes $A_1, A_2, ..., A_K$, where the points in each equivalence class share the same predecessor tree topology.[3] That is, the trees in an equivalence class are just shifted versions of one another; sharing the same pattern of axis parallel, fixed step size search offsets. Note that there can only be a finite number of distinct tree topologies given a bounded search depth.

For a given set of equivalent destination points, $A_k$, consider the shifted pre-images of $A_k$ corresponding to the n-odes of the tree, $A_{k1}, A_{k2}, ..., A_{kL}$. Note that these sets are disjoint, because the size of $A_k$'s containing cube, $A$, is smaller than the search step size $\epsilon$.

Now consider the measure that gets assigned to the destination set $A_k$ given the $\alpha$-weighted measures that are assigned to the initiating sets $A_{k\ell}$

$$\begin{aligned} \mu_j(A_k) &= \sum_{\ell=1}^{L} \mu_i(A_{k\ell}) \; \alpha_{k\ell} \\ &= \sum_{\ell=1}^{L} \mu_i(A_k) \; \alpha_{k\ell} \\ &\quad \text{since } \mu_i(A_{k\ell}) = \mu_i(A_k) \text{ by shift invariance} \\ &= \mu_i(A_k) \sum_{\ell=1}^{L} \alpha_{k\ell} \\ &= \mu_i(A_k) \\ &\quad \text{since } \sum_{\ell=1}^{L} \alpha_{k\ell} = 1 \text{ by construction.} \end{aligned}$$

Therefore, since $\mu_j(A_k) = \mu_i(A_k)$ for $k = 1, ..., K$ we have $\mu_j(A) = \mu_i(A)$. $\qquad \square$

From the lemma it quickly follows that $\mu_j(A) = \mu_i(A)$ for any Borel set $A$, and hence $\mu_j = \mu_i$.

Now note that the conditional measure on $X_i$ given $x_j$ is

$$\mu_{i|j}(x_i|x_j) = \begin{cases} \alpha_{ij} & \text{if } (x_i, x_j) \in B \\ 0 & \text{otherwise} \end{cases}$$

Thus, we have that for any joint event $A_i \times A_j$

$$\begin{aligned} \int_{A_i} \mu_{j|i}(A_j|x_i) \; \mu_i(dx_i) &= \mu_{ij}(A_i \times A_j) \\ &= \int_{A_j} \mu_{i|j}(A_i|x_j) \; \mu_j(dx_j) \end{aligned}$$

where $\mu_i$ and $\mu_j$ are both Lebesgue measure and $\mu_{j|i}$ and $\mu_{i|j}$ are as given above. This establishes what we require of the underlying measures.

Finally, returning to the original integration problem, by the (generalized) Fubini's theorem [Ash72, Theorem 2.6.4] we

---

[3]Technically we require each $A_k$ to be a Borel set, which imposes another benign technical restriction on the given task.

have that for a function $h(x_i, x_j)$ (again, satisfying broad technical conditions)

$$\int_{\mathbb{R}^n} \sum_{x_j \in B_i} h(x_i, x_j)\, \alpha_{ij}\, \mu_i(dx_i)$$

$$= \int_{\mathbb{R}^n} \int_{\mathbb{R}^n} h(x_i, x_j)\, \mu_{j|i}(dx_j | x_i)\, \mu_i(dx_i)$$
by definition of $\mu_{j|i}$

$$= \int_{\mathbb{R}^{2n}} h(x_i, x_j)\, \mu_{ij}(dx_i, dx_j)$$
by Fubini's theorem

$$= \int_{\mathbb{R}^n} \int_{\mathbb{R}^n} h(x_i, x_j)\, \mu_{i|j}(dx_i | x_j)\, \mu_j(dx_j)$$
by Fubini's theorem

$$= \int_{\mathbb{R}^n} \sum_{x_i \in B_j} h(x_i, x_j)\, \alpha_{ij}\, \mu_j(dx_j)$$
by definition of $\mu_{i|j}$

where both $\mu_i$ and $\mu_j$ are Lebesgue measure. This gives the desired result.

**Acknowledgements**

Research supported by NSERC and MITACS.